\documentclass[runningheads]{llncs}

\usepackage{eccv}

\usepackage{eccvabbrv}

\usepackage{graphicx}
\usepackage{booktabs}
\usepackage{amsmath}
\usepackage{amssymb}
\usepackage{pifont}
\usepackage{array}
\usepackage{comment}
\usepackage{epsfig}
\usepackage{url}
\usepackage{verbatim}
\usepackage{multirow}
\usepackage{colortbl}
\usepackage{bbding}
\usepackage{wasysym}
\usepackage[accsupp]{axessibility}  
\usepackage[rightcaption]{sidecap}
\usepackage{soul}
\usepackage{caption} 
\usepackage{wrapfig}
\usepackage{subcaption}
\usepackage{tabularx}

\usepackage{hyperref}

\usepackage{orcidlink}

\begin{document}

\title{Protecting NeRFs' Copyright via Plug-And-Play Watermarking Base Model} 
\titlerunning{NeRFProtector}



\author{Qi Song\inst{1,2},
Ziyuan Luo\inst{1,2},
Ka Chun Cheung\inst{2},
\\
Simon See\inst{2}
\and
Renjie Wan\inst{1}\thanks{Corresponding author. This work was done at Renjie’s Research Group at the
Department of Computer Science of Hong Kong Baptist University.}
}


\authorrunning{Q. Song et al.}

\institute{Department of Computer Science, Hong Kong Baptist University \and
NVIDIA AI Technology Center, NVIDIA
\\
\email{\{qisong,ziyuanluo\}@life.hkbu.edu.hk \{chcheung, ssee\}@nvidia.com renjiewan@hkbu.edu.hk }}

\maketitle

\begin{abstract}
Neural Radiance Fields (NeRFs) have become a key method for 3D scene representation. With the rising prominence and influence of NeRF, safeguarding its intellectual property has become increasingly important. In this paper, we propose \textbf{NeRFProtector}, which adopts a plug-and-play strategy to protect NeRF's copyright during its creation. NeRFProtector utilizes a pre-trained watermarking base model, enabling NeRF creators to embed binary messages directly while creating their NeRF. Our plug-and-play property ensures NeRF creators can flexibly choose NeRF variants without excessive modifications. Leveraging our newly designed progressive distillation, we demonstrate performance on par with several leading-edge neural rendering methods. Our project is available at: \url{https://qsong2001.github.io/NeRFProtector}. 
\keywords{ NeRF \and Copyright protection \and Plug-and-play}
\end{abstract}

\section{Introduction}
\label{sec:intro}

Neural Radiance Fields (NeRFs)~\cite{mildenhall2021nerf,chen2022tensorf,muller2022instant,fridovich2022plenoxels} mark a new way for 3D scene representation. As individuals might use radiance fields to represent their chosen 3D scenes and share these representations with the public, it becomes increasingly critical to establish a convenient copyright framework tailored to these novel representations during their creations. Then, once created, the copyright of these scene representations can be easily claimed in the event of an ownership breach.

While CopyRNeRF~\cite{luo2023copyrnerf} is developed to uphold copyright protection by embedding binary watermarks into NeRF models and then extracting watermarks from rendered images, its effectiveness is lessened by its weaknesses. \textbf{First}, a delay exists between creating the NeRF model and inserting ownership messages because binary messages are embedded into the NeRF model post-creation through model fine-tuning. Should malicious users acquire NeRF models in this delay, they could exploit them for nefarious intentions. \textbf{Second}, NeRF creators are required to jointly train a message extractor during message embedding, which makes the entire watermarking process exceedingly time-intensive and complex. NeRF creators or owners may abandon the watermarking due to its excessive complexity and difficulty. 

If NeRF creators are reluctant to use it, any fabulous designs for watermarking become meaningless. Therefore, in addition to the primary evaluation criteria for watermarking systems - robustness and invisibility of embedded binary messages~\cite{cox2002digitalsurvey} - we introduce NeRFProtector, a NeRF watermarking framework with a \textbf{plug-and-play} property. Such a plug-and-play property ensures that binary messages can be conveniently incorporated into existing NeRF variants with minimal modifications.

\begin{figure}[t]
  \centering
   \includegraphics[width=1\linewidth]{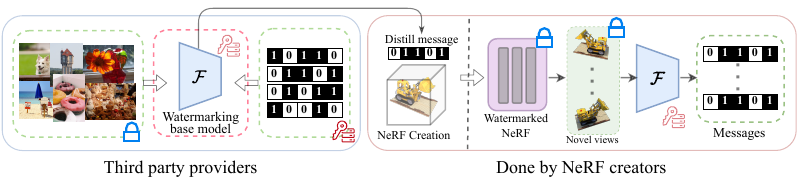}
   \caption{Proposed scenario of our method. NeRF creators can obtain a pre-trained watermarking extractor from a third party (\eg, open-source library) or train a message extractor separately via standard pipelines, as the \textbf{watermarking base model} $\mathcal{F}$. Once obtained, this base model is considered ``\textbf{plug-and-play}'' in our scenario. Throughout the NeRF creation process, creators can readily use this base model to embed watermarks in their NeRF. After the optimization of NeRF is complete, they obtain a watermarked  NeRF. When this watermarked NeRF is distributed publicly, the creators can then use the same base model to retrieve binary messages from newly rendered views, thereby asserting their ownership.}
   \label{fig:scenario}
\end{figure}

As shown in \cref{fig:scenario}, we use a watermarking base model to achieve the above goals. The watermarking base model can be obtained from a third party (\eg, some open-resource message extractors\footnote{\url{https://github.com/ando-khachatryan/HiDDeN}}) or trained separately via standard pipelines. NeRF creators simply need to select one watermarking base model and integrate it with their chosen NeRF variants (\eg, Instant-NGP~\cite{muller2022instant}, TensorRF~\cite{chen2022tensorf}, and so on), and then message embedding and NeRF optimization can proceed concurrently. Once optimization is complete, NeRF owners can use the base model to retrieve binary messages from the rendered contents of NeRF. This approach spares NeRF creators from extra efforts for modifying NeRF structures. Crucially, message embedding occurs during the creation of NeRF, leaving no window of opportunity for malicious attackers.  Based on our investigation, the message extractor from conventional image watermarking frameworks~\cite{zhu2018hidden,jia2021mbrs} can directly be our watermarking base model. These message extractors, through their training, have acquired knowledge of copyright messages and inherently possess the capability to retrieve such messages. Once NeRF creators acquire this watermarking base model, they can watermark their NeRF by distilling message knowledge from the extractor into their NeRF. Then, we only need to consider how to distill such established knowledge of binary messages into NeRF during creation.

Some methods like CLIP-NeRF~\cite{wang2022clipNerf} leverage knowledge from external CLIP model~\cite{radford2021learning_clip} to execute shape and color manipulation~\cite{cheng2024colorizing}. However, these methods necessitate modifications to NeRF's core structure for representation, which is inconsistent with our aim of ensuring watermarking compatibility with existing NeRF frameworks. To maintain simplicity, our proposed scenario opts for a simple Progressive Global Rendering (PGR) in place of NeRF's typical volume rendering. NeRF's volume rendering typically involves rendering a random subset of pixels, referred to as local rendering here, during optimization to manage computational demands. This limited pixel rendering is insufficient for embedding messages effectively within NeRFs. Rather than only rendering a small subset of pixels each time, our progressive global rendering renders all pixels across various image resolutions. This approach allows for more effective message embedding within NeRF's representation.

The framework of NeRFProtector is shown in \cref{fig:overview}. In our designs, we do not change the fundamental representation of NeRF. NeRF creators only need to use a pre-trained message extractor as the base model to finish both message embedding and extraction.  Despite the plug-and-play property of this base model, our framework can also satisfy the key criteria for robustness and invisibility.  Our key contribution can be concluded as follows: 

\begin{itemize}
\item A plug-and-play mode that NeRF creators easily watermark their NeRFs during the creation process, leaving no delays between NeRF creation and ownership message embedding. 

\item The utilization of a watermarking base model for efficient message embedding and extraction.

\item Progressive global rendering is proposed to effectively distill message knowledge from the watermarking base model into NeRFs by exploring relations between rendering strategies and watermark embedding.

\end{itemize}

The plug-and-play property of NeRFProtector ensures that minimal modifications are required to the architecture of NeRF representation, and the whole watermarking process can be conducted conveniently. Extensive experiments are conducted to verify the performance of our method, and the potential threats are also analyzed.

\section{Related work} 

\noindent{\textbf{Digital watermarking for 2D.}} 
Digital watermarking serves as a method for embedding copyright messages, referred to as watermarks, into various digital assets. These assets include images~\cite{zhu2018hidden,tancik2020stegastamp,li2021invisible_water,wang2024spy}, videos~\cite{luo2023dvmark_video,doerr2003guide_video,asikuzzaman2017overview_video}, and certain generative models~\cite{zhang2019steganogan,fernandez2023stable}. Researchers also investigate techniques that aim to ensure the watermark's robustness against various attacks~\cite{su2022robust,alzahrani2022enhanced_robust,evsutin2022watermarking_robust,praun1999robust} and common image processing operations such as compression, cropping, and resizing~\cite{zhu2018hidden,jia2021mbrs}.  The embedded copyright can be used for various purposes, including copyright protection and intellectual property protection. Our work leverages the prior knowledge of 2D watermarking and achieves copyright information protection for NeRF.

\noindent{\textbf{Neural Radiance Fields.}} Neural Radiance Fields (NeRF)~\cite{mildenhall2021nerf,muller2022instant,zhu2022neural,tang2024neural,zhu2023occlusion} is a method for 3D scene representation. It can represent a scene as a continuous 3D function that models the radiance at every point in space.
In recent years, NeRF has sparked a lot of research interest and has been widely adopted in the computer vision and graphics communities. NeRF has been extended to incorporate additional functionalities, such as text-to-3D~\cite{poole2022dreamfusion,lin2023magic3d,metzer2023latent}, image-to-3D~\cite{liu2023zero123,tang2023make3d,yu2021pixelnerf}, scene editing~\cite{liu2021editing_radiance,yuan2022nerf_edit}, and 3D model copyright protection~\cite{luo2023copyrnerf,li2023steganerf} tasks. These extensions further showcase the versatility and potential applications of NeRF in various domains. As the popularity and impact of NeRF continue to grow, the need for protecting its intellectual property becomes crucial.

\noindent{\textbf{Digital watermarking for 3D.}}
Traditional 3D watermarking approaches focus on  watermarking on polygonal meshes~\cite{ohbuchi2002frequency,praun1999robust,son2017perceptual} and point clouds~\cite{hamidi2019blind,liu2019novel} with explicit structures. A deep-learning-based approach~\cite{yoo2022deep} embeds binary messages in 3D meshes and extracts them from 2D rendered images.  For implicit 3D models such as NeRF, SteagaNeRF~\cite{li2023steganerf} introduces a technique for embedding data within NeRF. Specifically, CopyRNeRF~\cite{luo2023copyrnerf} suggests safeguarding NeRF's copyright by embedding binary messages into NeRF and subsequently extracting watermarks from the images rendered. CopyRNeRF~\cite{luo2023copyrnerf} also investigates several settings in the ownership claim of NeRF representation. However, CopyRNeRF~\cite{luo2023copyrnerf} is limited to embedding binary ownership messages through model fine-tuning post-creation of NeRF models, a process that potentially leaves room for exploitation by malicious individuals. Furthermore, its complex usage might discourage NeRF creators from adopting it. In contrast, our NeRFProtector enables the watermarking of NeRF models right at the point of creation, utilizing an easy-to-use plug-and-play watermarking base model.
\label{sec:related}

\section{Problem statement}
\label{sec:preliminaries}

\noindent{\textbf{NeRF.}} NeRF \cite{mildenhall2021nerf} takes in a 3D location $\mathbf{x} \in \mathbb{R}^3$ and viewing direction $\mathbf{d} \in \mathbb{R}^2$, then outputs a color value $\mathbf{c} \in \mathbb{R}^3$ and a density value $\sigma \in \mathbb{R}^+$. To produce a pixel’s color in a 2D image, samples of these values can be
composited along a ray according to volume rendering. Formally, $N_p$ points are sampled along a camera ray $r$ with color and density values $\{(\mathbf{c}_r^i, \sigma_r^i)\}_{i=1}^N$. The corresponding RGB color value $\hat{\mathbf{C}}(r)$ is obtained using alpha composition as
\begin{equation}
      \hat{\mathbf{C}}(r)=\sum_{i=1}^{N_p} T_r^i (1-\exp \left(-\sigma_r^i \delta_r^i\right)) \mathbf{c}_r^i, 
\end{equation}
where $T_r^i=\prod_{j=1}^{i-1}\left(\exp \left(-\sigma_r^i \delta_r^i\right)\right)$, and 
$\delta_r^i$ is the distance between adjacent sample points. The weights of NeRF can be optimized by minimizing a reconstruction loss between observations $\mathbf{C}$ and predictions $\hat{\mathbf{C}}$ as
\begin{equation}
  \mathcal{L}_{\text{content}} = \frac{1}{N_r}\sum_{m =1}^{N_r}\|\hat{\mathbf{C}}(r_m)-\mathbf{C}(r_m)\|_2^2,
  \label{eq:recon}
\end{equation}
where $N_r$ is the number of sampling pixels. 

\noindent{\textbf{Our scenario.}}  From the scenario shown in~\cref{fig:scenario}, NeRF creators can acquire the off-the-shelf watermarking base models from a third party or train a base model separately via standard pipelines. Then, they can directly combine the base model with their NeRF optimization. Once the optimization has concluded, the message embedding also concludes. Then, the base model can be further used to extract binary messages for ownership claims.

Our approach diverges from conventional watermarking techniques~\cite{luo2023copyrnerf} for NeRF models. Rather than embedding the ownership messages after the creation of NeRF, we ensure NeRF models are protected right from their outset, substantially narrowing the window of opportunity for malicious actors. Additionally, this method permits NeRF creators to train their models with minimal deviation from standard procedures, avoiding the extra workload of training supplementary modules.

\section{Proposed NeRFProtector}
\label{sec:method}
\begin{figure*}[t]
  \centering
   \includegraphics[width=1.0\linewidth]{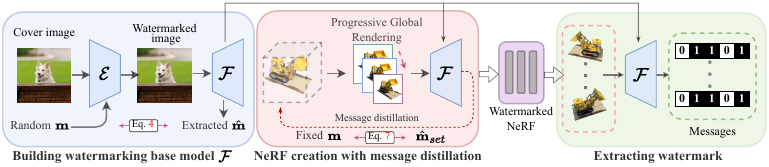}
   \caption{Our \textbf{plug-and-play} method to watermark NeRF during its creation. \textbf{(1) Building watermarking base model}: The watermarking base model $\mathcal{F}$ can be sourced from a third party (\cref{sec:pre-training}). We implement a HiDDeN~\cite{zhu2018hidden} framework to get the pre-trained message extractor as our watermarking base model $\mathcal{F}$. During the training, the encoder $\mathcal{E}$ encodes a randomly selected 48-kit message $\mathbf{m}$ and cover image $\mathbf{I}_o$ and outputs a watermarked image $\mathbf{I}_{en}$. Then the message extractor $\mathcal{F}$ extracts embedded message $\hat{\mathbf{m}}$ from the watermarked image. Message encoder $\mathcal{E}$ is discarded after building watermarking base model $\mathcal{F}$.  \textbf{(2) NeRF creation with message distillation}: NeRF creators first fix a copyright message $\mathbf{m}$, then employ this base model $\mathcal{F}$ to embed selected watermarks to NeRFs during the creation process via Progressive Global Rendering (PGR) and message distillation (\cref{sec:distilling}). When the optimization of NeRF is finalized, creators immediately obtain a watermarked NeRF. \textbf{(3) Extracting watermark}: Subsequently, they can utilize the base model $\mathcal{F}$ to retrieve binary watermarks from the rendered images, asserting their ownership.}
   \label{fig:overview}
\end{figure*}

Our NeRFProtector aims at embedding \textit{k}-bit binary messages $\mathbf{m} \in \{0,1\}^k$ during the creation of NeRF. As our scenario relies on an important base model for watermarking, we first introduce it in \cref{sec:pre-training}. Then, we introduce how to distill knowledge from this base model into NeRF via a newly designed progressive distillation in~\cref{sec:distilling}. 

\subsection{Building watermarking base model}
\label{sec:pre-training}
The essence of our approach is to switch out the current processes of message embedding and extraction with a watermarking base model. We envision this base model to possess a \textbf{plug-and-play} property, enabling NeRF creators to seamlessly incorporate it into their NeRFs with only minor modifications to established architecture. Additionally, this base model is designed to withstand common image-level distortions, ensuring robust message extraction even from distorted rendered images.

For simplicity, we choose the extractor from HiDDeN ~\cite{zhu2018hidden}, a well-established deep-learning-based watermarking framework, to serve as our base model. As a message extractor for image watermarking, the base model naturally owns the capability to extract binary messages from watermarked images. Besides, we can easily make the extraction more robust by employing established studies in image watermarking~\cite{zhu2018hidden,jia2021mbrs}. At last, the base model has learned the message pattern during its training, and we can then distill such learned knowledge into the representation of NeRF during its optimization.

Due to the requirement for plug-and-play, we refrain from making additional modifications to HiDDeN~\cite{zhu2018hidden}. 
In general, HiDDeN~\cite{zhu2018hidden} jointly optimizes the watermark encoder $\mathcal{E}$ and the watermarking base $\mathcal{F}$ to robustly embed \textit{k}-bit binary messages into images. 
Specifically, during optimization, the encoder $\mathcal{E}$ takes a cover image $\mathbf{I}_o$ and a message $\mathbf{m} \in \{0,1\}^k$ as inputs and output a watermarked image $\mathbf{I}_{en}$ as follows, 
\begin{equation}
    \mathbf{I}_{en} = \mathcal{E}(\mathbf{I}_o,\mathbf{m}).
    \label{eq:encoder}
\end{equation}

Binary message $\hat{\mathbf{m}}$ is then extracted from distorted images using a watermark extraction network $\mathcal{F}$ as $\hat{\mathbf{m}} = \mathcal{F}(T(\mathbf{I}_{en}))$. Then, the whole network is optimized by minimizing the Binary Cross Entropy (BCE) loss between the original binary message $\mathbf{m}$ and extracted binary message $\hat{\mathbf{m}}$ as:
\begin{equation}
\mathcal{L}_\text{m} = \text{BCE} (\mathbf{m},\hat{\mathbf{m}}),
\label{eq:bce}
\end{equation}
where $ \text{BCE}(\mathbf{m},\hat{\mathbf{m}}) = -\left(\mathbf{m} \cdot \log(\tau(\hat{\mathbf{m}})) + (1-\mathbf{m}) \cdot \log(1-\tau(\hat{\mathbf{m}})\right)$ and $\tau$ is a sigmoid operation. After the training, we discard the encoder $\mathcal{E}$ in \cref{eq:encoder} and only use the extractor $\mathcal{F}$ as the watermarking base model.  To improve the robustness, we follow HiDDeN~\cite{zhu2018hidden} and employ a randomly selected transformation layer $T$ during training to ensure that the hidden messages are robust to common image distortions. 


\subsection{Distilling watermarks into NeRF progressively}
\label{sec:distilling}

The base model has encapsulated all necessary knowledge about watermarking. We further distill the knowledge patterns into NeRF as they are being created. Then, the binary messages can be robustly extracted from rendered images regardless of the viewpoints. Although various approaches exist for distilling external knowledge into the representations of NeRF~\cite{fan2022nerf-sos,wang2022r2l}, they necessitate changes to the essential structures of NeRF, conflicting with our need for \textbf{a plug-and-play property}. We achieve the watermark distillation by only making alternations to the rendering scheme of NeRF, the key obstacles for our distillation.

As displayed in \cref{eq:recon}, NeRF hinges on a random and small subset of pixels ($N_r$ pixels in \cref{eq:recon}) for rendering. Despite its efficiency in preventing the computational demand from being too high, the message patterns can only be embedded into random positions during distillation if only such local rendering is used. Then, as shown in \cref{fig:abl}, when the optimization settles down, the message pattern cannot effectively form meaningful patterns that can be identified and extracted by the base model. 
As shown in  \cref{fig:overview}, we substitute such conventional local rendering in established NeRF with a progressive global rendering that is capable of rendering all pixels across various image resolutions. This modification does not alter the representation of NeRF, ensuring that our mechanism remains adaptable and can be easily applied to a range of NeRF variants~\cite{fridovich2022plenoxels,muller2022instant,chen2022tensorf}. 

\noindent {\textbf{Progressive Global Rendering (PGR).}} For our PGR, as displayed in \cref{fig:overview}, rather than only sampling a random subset of pixels, during rendering, we sample rays corresponding to all pixels across various scales to gain a comprehensive understanding of global information during optimization. By embedding messages on a global scale, they become deeply integrated into the scene's representation. This ensures the consistent properties of binary watermarks from various viewpoints, allowing message extraction irrespective of rendering viewpoints. Additionally, since 3D information often exhibits distinct characteristics in 2D projections at varying scales~\cite{qi2017pointnet++}, progressive rendering accentuates these resolution-dependent properties, aiding in message distillation. Furthermore, the globally rendered images are all with reduced resolution. Then, our PGR can achieve comparable results without necessitating full-resolution images for optimization and make the computational cost acceptable.

As shown in \cref{fig:overview}, we consider rendering $N_k$ progressive views $\hat{\mathbf{I}}_{set}$ at different scales denoted as follows:
\begin{equation}
    \hat{\mathbf{I}}_{set} = \{\hat{\mathbf{I}}_n\}_ {n=1}^{N_k} = \{\hat{\mathbf{I}}_1,\hat{\mathbf{I}}_2, ... ,\hat{\mathbf{I}}_{N_k} \} ,
\end{equation}
where $N_k$ and $\hat{\mathbf{I}}_{n}$ denote the number of layers and the cascade rendered view at different scales. Each $\hat{\mathbf{I}}_n  \in \mathbb{R}^{\frac{W}{2^n} \times \frac{H}{2^n} \times 3}$ from progressive views set $\hat{\mathbf{I}}_{set}$ is obtained by rendering global rays of at corresponding resolution. In our setting, we empirically set $N_k = 3$, a value that can well balance efficiency and performance.

\noindent{\textbf{Distilling messages.}} We then utilize the base model $\mathcal{F}$ and progressive global rendering to distill message patterns into NeRF. For $N_k$ rendered cascade views $\hat{\mathbf{I}}_{set}$ at different scales, we use the watermarking base model $\mathcal{F}$ to extract the embedded message $\hat{\mathbf{m}}_{set}$ as follows:
\begin{equation}
\hat{\mathbf{m}}_{set} = \mathcal{F}(\hat{\mathbf{I}}_{set}) \\
= \{\hat{\mathbf{m}}_1,\hat{\mathbf{m}}_2,...,\hat{\mathbf{m}}_{N_k} \}. 
\end{equation}
Then a distillation loss $\mathcal{L}_{\text{dis}}$ is computed by calculating the BCE loss between the predicted message set $\hat{\mathbf{m}}_{set}$ and the ground truth message $\mathbf{m}$:
\begin{equation}
\mathcal{L}_{\text{dis}} = \sum_{i=1}^{N_k} \alpha_i \cdot \text{BCE}(\mathbf{m}, \hat{\mathbf{m}}_i),
\end{equation}
where $N_k$ is the number of cascade layers, and $\alpha_i$ represents the weight of $i$ layer's loss function. During the creation, minimizing this equation can distill messages into NeRF. 

To make the distilled message pattern invisible, we enforce the high similarity between the rendering image and its corresponding ground truth as follows: 
\begin{equation}
  \mathcal{L}_{\text{inv}} = \|\hat{\mathbf{I}}_{1}-\mathbf{I}_1\|_2^2,
  \label{eq:global}
\end{equation}
where $\hat{\mathbf{I}}_{1}$ denotes the cascade rendered images with the highest resolution, and $\mathbf{I}_1$ is its corresponding ground truth. The ground truth is with a reduced resolution to be compatible with $\hat{\mathbf{I}}_{1}$.

Besides progressive global rendering, we still use local rendering for building radiance fields. During optimization, the local rendering randomly selects $N_r$ pixels for rendering, which ensures the local pattern can be effectively considered during the message distillation. Thus, our whole loss functions for creating NeRF with a plug-and-play watermarking base model can be concluded as follows: 

\begin{equation}
\mathcal{L}=\lambda_1\mathcal{L}_{\text{local}}+\lambda_2\mathcal{L}_{\text{inv}}+\lambda_3\mathcal{L}_{\text{dis}}, 
\label{equ:all_loss}
\end{equation}
where $\lambda_1$, $\lambda_2$, and $\lambda_3$ are hyperparameters that balance the performance between embedding watermarking and reconstruction, and the $\mathcal{L}_{\text{local}}$ is same to \cref{eq:recon}. The network setup described above can be effectively adapted for embedding copyright messages into NeRF. PGR is only used during the optimization. Once NeRF is created, users can still render images via established ways~\cite{mildenhall2021nerf}.

\subsection{Implementation details}
\label{implementation}
 
\underline{\textit{For NeRF}}, we implement our method with Instant-NGP~\cite{muller2022instant}. Instant-NGP takes a 3D grid as input and encodes it using a grid encoder. It then predicts density values $\sigma$ using two-layer MLP with 64 channels. For the color branch,  a three-layer MLP with 64 channels is used to predict color $c$ and $\sigma$ from view direction $(\theta,\phi)$. As for the rendering, the number of cascade layers $N_k$ is set to 3, and the rendered pixel count $N_r$ for the local rendering is set to $4096$, respectively. 
\underline{\textit{For base model}}, we implement our method with HiDDeN~\cite{zhu2018hidden}. The watermark encoder $\mathcal{E}$ and extractor $\mathcal{F}$ in HiDDeN~\cite{zhu2018hidden} are jointly trained on COCO~\cite{lin2014microsoft} dataset.
During its training process, encoder $\mathcal{E}$ encodes a randomly selected 48-kit message $\mathbf{m}$ and cover image $\mathbf{I}_o$ and predicts a watermarked image $\mathbf{I}_{en}$. Then extractor $\mathcal{F}$ extracts
embedded messages from watermarked image $\mathbf{I}_{en}$.
The whole network is optimized with Adam~\cite{kingma2014adam} optimizer in $100K$ iterations.

During NeRF creation, the NeRF creator chooses one preferable message $\mathbf{m}$ as copyright messages. We fix the weight of watermarking base model $\mathcal{F}$ and minimize the distance between the chosen message $\mathbf{m}$ and extracted message $\hat{\mathbf{m}}_{set}$.  Hyperparameters in \cref{equ:all_loss} are set as $\lambda_1 = 0.01$ and $\lambda_3 = 0.001$. We use the Adam~\cite{kingma2014adam} optimizer with setting the values of $\beta_1 = 0.9$, $\beta_2 = 0.999$, and a learning rate of 0.01. The learning rate decays following the exponential scheduler during optimization. We train our model for each scene on a single NVIDIA A100 GPU. 


\section{Experiments}
\label{sec:experiment}

\subsection{Experiment settings}
\label{sec:settings}

\noindent{\textbf{Dataset.}} 
We conduct experiments on the Blender~\cite{mildenhall2021nerf} and LLFF~\cite{mildenhall2019local_llff} dataset. The Blender dataset comprises 8 synthetic object images captured from different viewpoints, while the LLFF dataset consists of 8 real-world scenes captured in a forward-facing manner. We choose these datasets based on previous research~\cite{mildenhall2021nerf,muller2022instant} and use them to assess the bit accuracy and reconstruction performance of NeRF. For each scene, we render testing views and calculate their average values for our analysis.

\noindent{\textbf{Baselines.}} We compare four strategies to guarantee a fair comparison: 1) HiDDeN~\cite{zhu2018hidden} + NeRF~\cite{muller2022instant}: processing images with classical 2D watermarking method HiDDeN~\cite{zhu2018hidden}  before training the NeRF model; 2) MBRS ~\cite{jia2021mbrs} + NeRF~\cite{muller2022instant}: processing images with a recently proposed 2D watermarking method MBRS~\cite{jia2021mbrs} before training the NeRF model; 3) CopyRNeRF~\cite{luo2023copyrnerf}: a state-of-the-art method for protecting the copyright of NeRF models via digital watermarking. To evaluate the impact of watermarks on reconstruction, we also compare the results with the non-watermarked version of NeRF; 4) NeRF~\cite{muller2022instant} w/o watermark: we also compare with NeRF~\cite{muller2022instant} w/o watermark to evaluate the reconstruction quality.
\textbf{To further illustrate our plug-and-play properties, we implement our method with other NeRF variants~\cite{fridovich2022plenoxels,chen2022tensorf} and base model~\cite{evsutin2022watermarking_robust} as demonstrated in \cref{ablation_study}}. 

\begin{figure*}[th]
  \centering
   \includegraphics[width=1.0\linewidth]{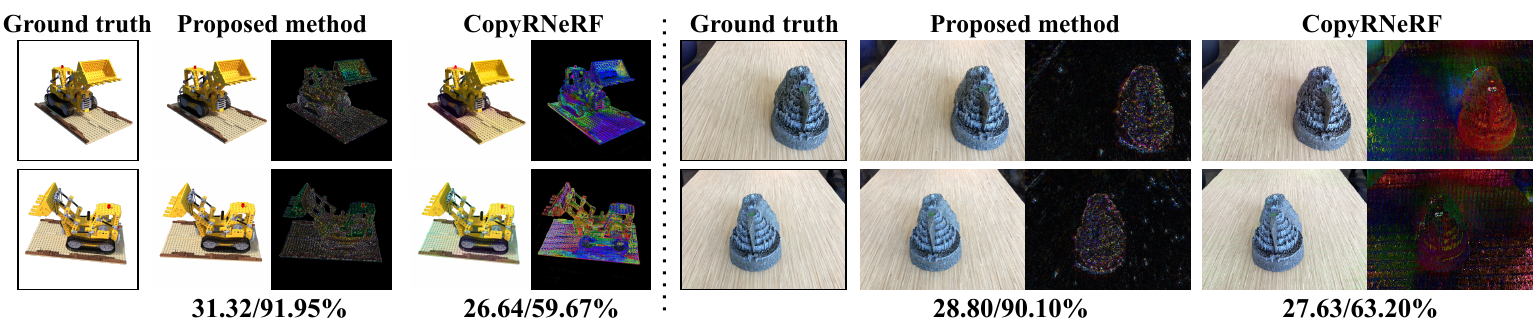}
   \caption{Visual quality comparisons with CopyRNeRF~\cite{luo2023copyrnerf}. The texts below the images show the results of PSNR and bit accuracy. We show the differences $ \times 10$ between the rendered views and the ground truth for both methods. Our NeRFProtector exhibits better consistency across multiple viewpoints, achieving a well-balanced trade-off between reconstruction quality and bit accuracy.
}
   \label{fig:main_visual}
\end{figure*}

\noindent{\textbf{Evaluation methodology.}} We assess our method against other baselines using the established digital watermarking standards of invisibility and robustness~\cite{cox2002digitalsurvey}. For \textbf{invisibility}, we measure the performance by employing metrics including PSNR, SSIM~\cite{wang2004image}, and LPIPS~\cite{zhang2018unreasonable} to compare the visual fidelity of the output images after binary message embedding. For \textbf{robustness}, we conduct experiments to determine if the embedded messages can be accurately extracted by assessing the bit accuracy under different types of image distortions. We consider various distortions~\cite{fernandez2023stable,qi2023degradation} for message extraction, including JPEG compression, image crop, image scaling, contrast changes, and text overlay. We directly set the bit length used in our experiments as $48$ bits, the largest length used in previous methods for watermarking 3D models~\cite{luo2023copyrnerf,yoo2022deep}. We further discuss potential intended watermarking removal technologies in \cref{threat_model}. 

\begin{table*}[tbp]
\centering
\caption{Quantitative results of visual representation and watermarking extraction performance. We report the results of watermarking robustness when rendered images meet common distortions (\eg, Crop, Resize, and JPEG compression). ``None'' indicates that no distortion has been applied.  ``N/A'' denotes \ul{Not Applicable} as NeRF w/o watermark does not involve message embedding.}
\label{tab:main_table}%
\resizebox{0.98\textwidth}{!}{%
\begin{tabular}{c|c|ccc|cccc}
\toprule
\multirow{2}{*}{Dataset} & \multirow{2}{*}{Method}  & \multirow{2}{*}{PSNR $\uparrow$} & \multirow{2}{*}{SSIM $\uparrow$} & \multirow{2}{*}{LPIPS $\downarrow$} &\multicolumn{4}{c}{Bit accuracy \% }  \\
&  & & &  &None   &Crop   &Resize &JPEG \\
\cline{6-7}
\hline
\multirow{5}{*}{Blender-test} 
&\cellcolor{red!10}NeRF w/o watermark &\cellcolor{red!10}30.62  &\cellcolor{red!10}0.9579 &\cellcolor{red!10}0.0343 &\multicolumn{4}{c}{\cellcolor{red!10}N/A}\\
\cline{2-9}
& HiDDeN~\cite{zhu2018hidden}+NeRF  &28.30 &0.9219  &0.0495  &50.29 &50.03 &52.32 &50.84 \\
& MBRS~\cite{jia2021mbrs}+NeRF  &24.17 &0.8461 &0.2967 &50.53 &49.64 &51.72 &49.81\\
& CopyRNeRF~\cite{luo2023copyrnerf}  & 25.50 &0.9073	&0.0885   &62.15 &56.63 &57.32 &58.41\\ 
&\cellcolor{cyan!10}\textbf{NeRFProtector} &\cellcolor{cyan!10}\textbf{29.26} &\cellcolor{cyan!10}\textbf{0.9393} &\cellcolor{cyan!10}\textbf{0.0483} &\cellcolor{cyan!10}\textbf{92.69} &\cellcolor{cyan!10}\textbf{92.95}  &\cellcolor{cyan!10}\textbf{91.87} &\cellcolor{cyan!10}\textbf{78.62 }\\
\hline 
\multirow{5}{*}{LLFF-test} 
&\cellcolor{red!10}NeRF w/o watermark  &\cellcolor{red!10}26.37 &\cellcolor{red!10}0.8352	&\cellcolor{red!10}0.1013	 &\multicolumn{4}{c}{\cellcolor{red!10}N/A}\\
\cline{2-9}
& HiDDeN~\cite{zhu2018hidden}+NeRF &25.70 &0.8300 &0.1096 &51.69 &50.48 &50.53 &51.11\\
& MBRS~\cite{jia2021mbrs}+NeRF  &25.44 &0.8198	&0.1238 &50.84 &51.55 &50.85 &49.08\\
& CopyRNeRF~\cite{luo2023copyrnerf} &25.80 &0.8302 & 0.1035  &63.72 &60.45 &55.34 &54.11\\ 
&\cellcolor{cyan!10} \textbf{NeRFProtector} &\cellcolor{cyan!10}\textbf{26.82} &\cellcolor{cyan!10}\textbf{0.8569}	&\cellcolor{cyan!10}\textbf{0.0834}  &\cellcolor{cyan!10}\textbf{96.99} &\cellcolor{cyan!10}\textbf{93.57} &\cellcolor{cyan!10}\textbf{80.53} &\cellcolor{cyan!10}\textbf{76.26}  \\
\bottomrule
\end{tabular}
}

\end{table*}

\noindent{\textbf{Invisibility against Bit accuracy.}} As shown in \cref{tab:main_table} and \cref{fig:main_visual}, we first investigate the invisibility of our watermarks. This is to show whether the embedded messages undermine the visual quality. Similar to previous studies~\cite{luo2023copyrnerf}, for HiDDeN~\cite{zhu2018hidden}/MBRS~\cite{jia2021mbrs} + NeRF, binary messages cannot be effectively extracted from rendered images, though they all achieve acceptable visual quality. CopyRNeRF~\cite{luo2023copyrnerf} achieves the second-best results while it is still largely below our values for bit accuracy. Only our approach can effectively balance visual quality (PSNR/SSIM/LPIPS) and message embedding (Bit accuracy). Besides, from \cref{fig:main_visual}, the viewpoints have a higher impact on the bit accuracy of message extraction, while our method can achieve more consistent performances across different viewpoints.

\begin{figure}
  \centering
   \includegraphics[width=0.50\linewidth]{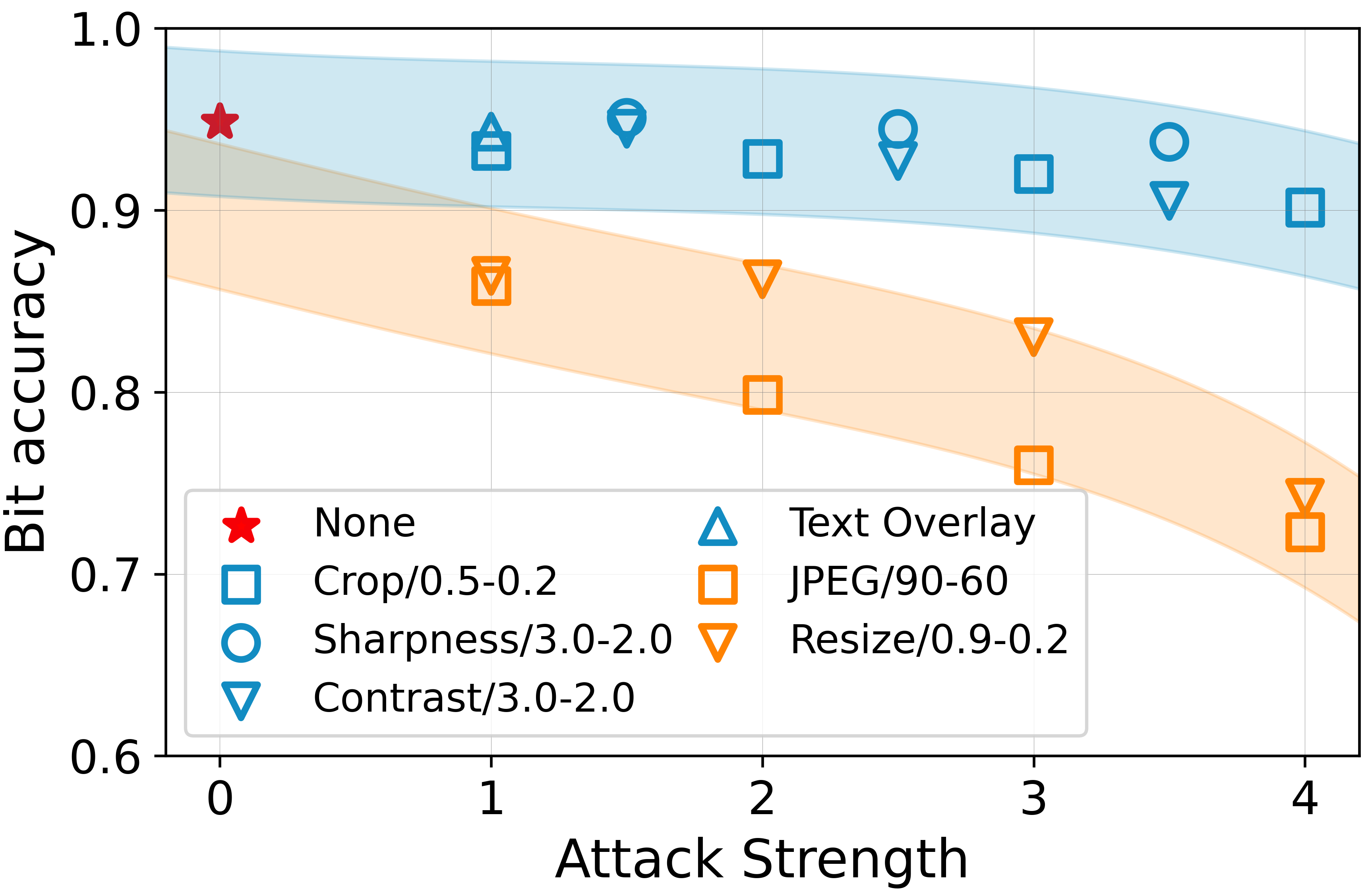}
   \caption{Image-level distortion experiment on common image distortions on rendered views to determine if the extractor can still extract the message. Operations like sharpness, cropping, contrast, and text overlay have little impact on message extraction. Even after severe operations (JPEG and Resize), our method still maintains satisfactory accuracy ($\geq 70\%$).}
   \label{fig:exp_attack}
\end{figure}

\noindent{\textbf{Is it robust to common distortion?}} Our method is robust to common image-level distortions since we have considered their impact during building watermarking base model $\mathcal{F}$. As shown in \cref{tab:main_table} and \ref{fig:exp_attack}, when image operations are applied to the rendered image, our base model can still effectively extract the embedded binary messages on both evaluation datasets.

\noindent{\textbf{Effectiveness.}} We compare the training time of our method and CopyRNeRF~\cite{luo2023copyrnerf} with same message length setting. In contrast to CopyRNeRF, our approach costs only around $50$m. As CopyRNeRF~\cite{luo2023copyrnerf} jointly trains NeRF and the watermarking extractor at each step with a random message, it significantly extends the time required to embed the watermark ($30$h).

\subsection{Ablation study}
\label{ablation_study}
\noindent{\textbf{Is progressive global rendering good?}} We propose a progressive distillation to effectively distill message patterns from the base model to NeRF. We compare three settings in \cref{tab:abl}:  1) \textbf{Local rendering only}, the rendering strategy used in vanilla NeRF; 2) \textbf{Single-scale global rendering}: we replace the progressive global rendering with a single-scale rendering; 3) \textbf{Our complete progressive global rendering}. 

\begin{wraptable}[10]{r}{0.5\linewidth}
  \centering
  \vspace{-1.65cm}
\caption{Quantitative results for evaluating progressive global rendering. Local denotes the rendering strategy used in vanilla NeRF. Single-scale represents single-scale global rendering from progressive rendering. Progressive denotes our proposed progressive global rendering.}
  \label{tab:abl}%
  \resizebox{1.0\linewidth}{!}{
    \begin{tabular}{l|cccc}
    \toprule
     \multirow{2}{*}{Settings} & \multirow{2}{*}{PSNR $\uparrow$} & \multirow{2}{*}{SSIM $\uparrow$} & \multirow{2}{*}{LPIPS $\downarrow$} &\multirow{2}{*}{Bit acc. \%} \\
       & & & & \\
    \hline
    \textcircled{\small 1} Local  &30.38	&0.9521	&0.0360	&45.99 \\
    \hline
    \textcircled{\small 2} Single-scale& \multirow{1}{*}{29.57}       &\multirow{1}{*}{0.9402}  &\multirow{1}{*}{0.0449}  &\multirow{1}{*}{87.27}  \\
    \hline
    \textcircled{\small 3} Progressive &29.26      &\multirow{1}{*}{0.9394}       &\multirow{1}{*}{0.0483}  &  \multirow{1}{*}{92.69} \\
    \bottomrule
    \end{tabular}%
    }
\end{wraptable}

\begin{wrapfigure}[16]{r}{0.4\linewidth}
  \centering
   \includegraphics[width=1.0\linewidth]{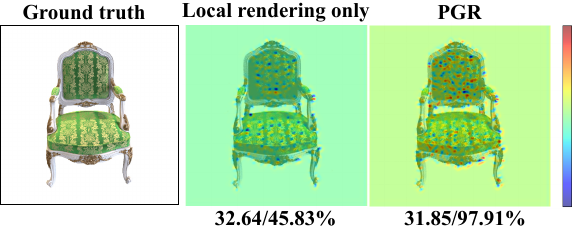}
   \caption{Visual heat map comparisons of local rendering and progressive global rendering. The text below denotes PSNR/Bit accuracy. The colors represent the area's impact on message extraction. Our PGR can distill messages into NeRF with high accuracy.
}
   \label{fig:abl}
\end{wrapfigure}

As shown in \cref{tab:abl}, the local rendering in current mechanisms of NeRF cannot guarantee successful message embedding, where the bit accuracy is kept at a very low level. Besides, we also show the heat maps obtained from our progressive global rendering and classical local rendering in \cref{fig:abl}. From the results for ``Single-scale global rendering'', even with a single-scale global rendering, the accuracy of embedding bits is further enhanced while ensuring the reconstruction quality. Our experiments have demonstrated that the progressive rendering approach more effectively incorporates distillation information into the NeRF model. 

\noindent{\textbf{Can proposed method adapt to other NeRF variants?}} Our base model for watermarking can also be easily adapted to other NeRF variants. As displayed in \cref{tab:backbone}, besides Instant-NGP~\cite{muller2022instant} used in our experiments, we further combine the base model with TensorRF~\cite{chen2022tensorf}, and Plenoxels~\cite{fridovich2022plenoxels}. From the experimental results, our base model for watermarking can also achieve comparable performance on different NeRF variants. 

\noindent{\textbf{Can the base model be other extractors?}} In our default setting, we use the extractor of HiDDeN~\cite{zhu2018hidden} as our base model. However, we can also consider using other extractors trained in different settings~\cite{jia2021mbrs} as our base model.  In \cref{tab:extractor}, we show that message extractor from MBRS~\cite{jia2021mbrs} can also be used as the base model, and they also achieve comparable performance.

\begin{table}[h]
\centering
\begin{minipage}{0.50\textwidth}
\caption{Quantitative results on combining the same watermarking base model with different NeRF variants. The extractor of HiDDeN~\cite{zhu2018hidden} is employed as the watermarking base model $\mathcal{F}_1$. We combine $\mathcal{F}_1$ with Instant-NGP~\cite{muller2022instant}, TensorRF~\cite{chen2022tensorf}, and Plenoxels~\cite{fridovich2022plenoxels} to see whether the base model can be adapted to NeRF variants in a plug-and-play manner. N/A denotes not applicable as message embedding is not involved.}
\label{tab:backbone}
\centering
  \resizebox{0.9\linewidth}{!}{
    \begin{tabular}{c|cccc}
    \toprule
    Method & PSNR $\uparrow$ &SSIM $\uparrow$ &LPIPS $\downarrow$ & Bit acc.\% \\
    \hline
    \cellcolor{red!10}Instant NGP~\cite{muller2022instant} &\cellcolor{red!10}33.42	&\cellcolor{red!10}0.9709	 &\cellcolor{red!10}0.0157  &\cellcolor{red!10}N/A \\
    Instant NGP + $\mathcal{F}_1$  &32.92 &0.9697	&0.0161	&91.96 \\
    \hline
    \cellcolor{red!10}TensorRF~\cite{chen2022tensorf} &\cellcolor{red!10}34.96 &\cellcolor{red!10}0.9786 &\cellcolor{red!10}0.0105  & \cellcolor{red!10}N/A \\
    TensorRF + $\mathcal{F}_1$& 32.73 & 0.9743& 0.0125  & 89.35 \\
    \hline
\cellcolor{red!10}Plenoxels~\cite{fridovich2022plenoxels}  &\cellcolor{red!10}35.80 &\cellcolor{red!10}0.9872 &\cellcolor{red!10}0.0102  &\cellcolor{red!10}N/A \\
    Plenoxels + $\mathcal{F}_1$ &34.19 &0.9730 &0.0127 &97.92 \\
    \bottomrule
    \end{tabular}
}
\end{minipage}
\hfill
\begin{minipage}{0.45\textwidth}
\caption{Quantitative results on combining different watermarking base model. $\mathcal{F}_1$ and $\mathcal{F}_2$ denote HiDDeN~\cite{zhu2018hidden} and MBRS~\cite{jia2021mbrs} extractors. We combine a different base model from MBRS~\cite{jia2021mbrs} with Instant-NGP~\cite{muller2022instant} to see whether the backbone of NeRF can be adapted to other base models. N/A denotes not applicable as message embedding is not involved.}
\label{tab:extractor}
\centering
  \resizebox{1.0\linewidth}{!}{
    \begin{tabular}{c|cccc}
    \toprule
    Method & PSNR $\uparrow$ &SSIM $\uparrow$ &LPIPS $\downarrow$ & Bit acc.\% \\
    \hline
    \cellcolor{red!10}Instant-NGP~\cite{muller2022instant} &\cellcolor{red!10}33.42	&\cellcolor{red!10}0.9709	 &\cellcolor{red!10}0.0157  &\cellcolor{red!10}N/A \\
    Instant-NGP + $\mathcal{F}_1$  &32.92 &0.9697	&0.0161	&91.96 \\
    Instant-NGP + $\mathcal{F}_2$  &31.71 &0.9635	&0.0152	&89.13 \\
    \bottomrule
    \end{tabular}
}
\end{minipage}
\end{table}

\subsection{Analyzing potential threats} 
\label{threat_model}

Our target is to propose a framework that is able to utilize the robustness within the base model to defend some common image-level threats. The experiments also demonstrate the advantages of our approach. We further examine the embedded watermark’s robustness to more potential intentional tampering and threats.

\begin{figure}[h]
    \centering
    \begin{minipage}{0.58\textwidth}
        \centering
        \includegraphics[width=0.75\linewidth]{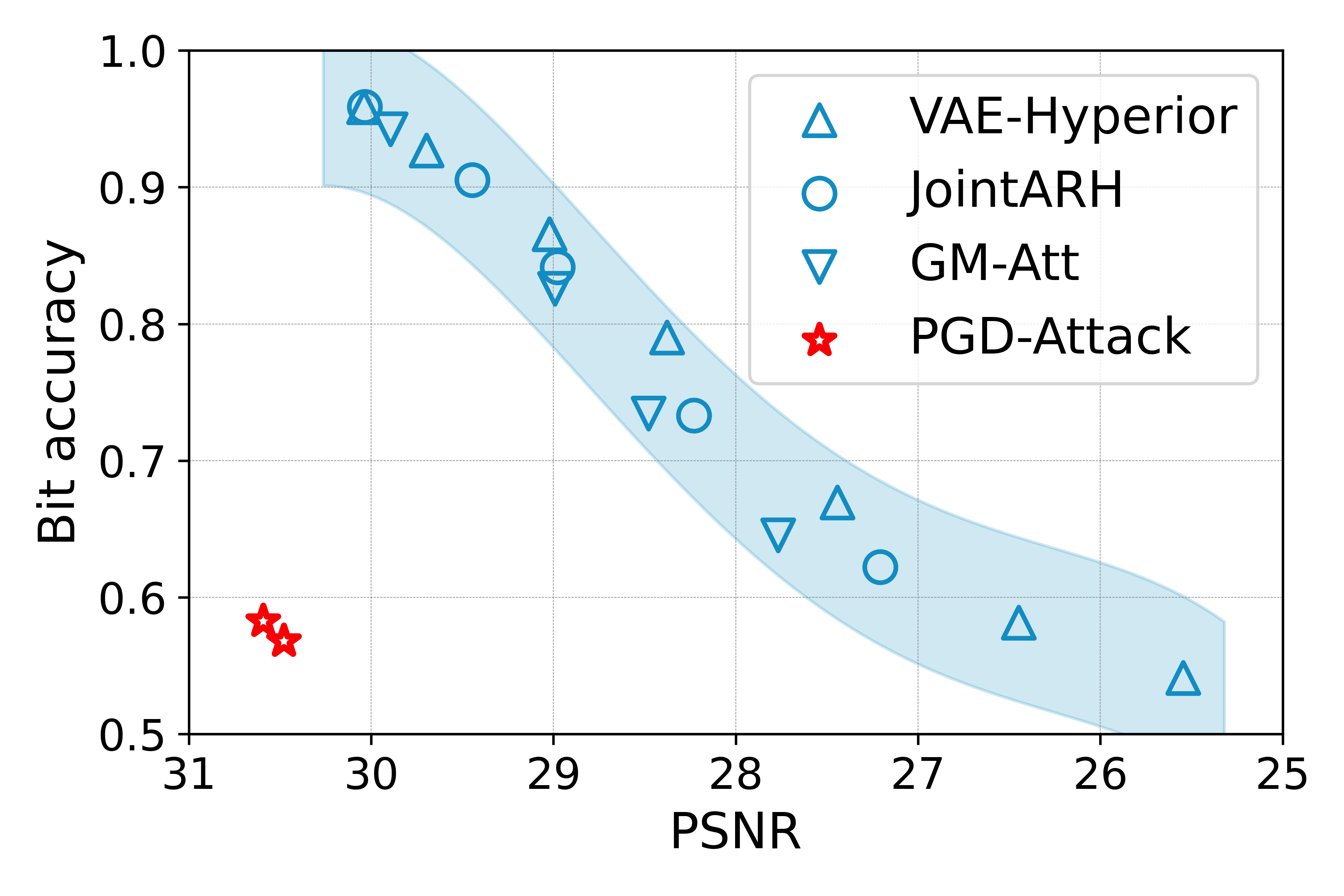}
       \caption{Attacks on rendered images. We evaluate neural compression methods including VAE-Hyperior~\cite{balle2018variational_bmsh2018}, JointARH~\cite{minnen2018joint} and GM-Att~\cite{cheng2020learned}. PGD-attack~\cite{madry2017towards} is also applied to attack the base model. Attacks using image compression methods can only succeed when there is a significant decrease in image quality. However, it is possible to remove watermarks with PGD attack~\cite{madry2017towards} on the base model without obvious distortion.} 
       \label{fig:image_level_attack}
    \end{minipage}
    \hfill
    \begin{minipage}{0.38\textwidth}
        \centering
  \includegraphics[width=1.0\linewidth]{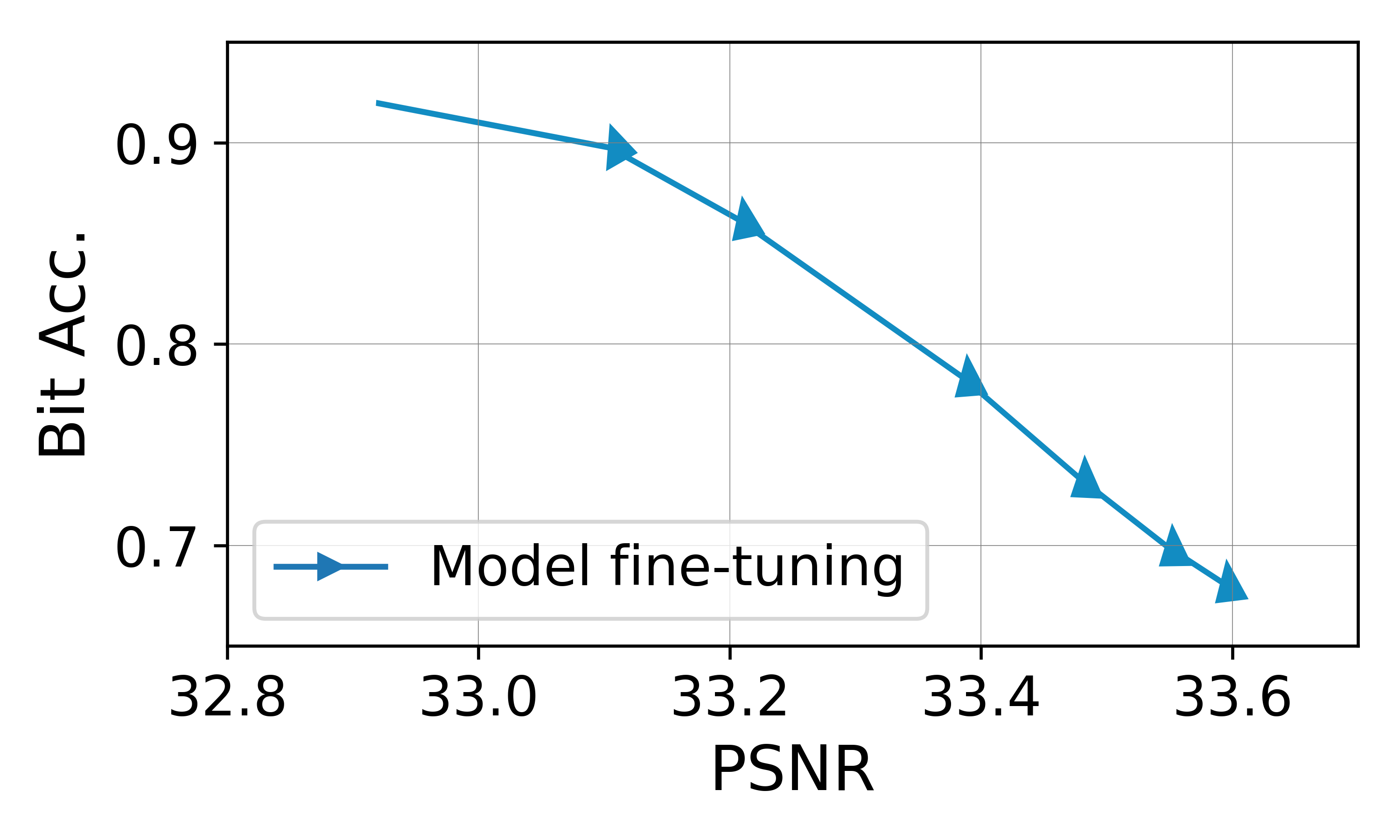}
  \caption{In an extreme situation, attackers might gain access to the images utilized in creating NeRF. With these unwatermarked images, they could then fine-tune the model to eliminate the watermarks. In this example, as the fine-tuning progresses, the PSNR keeps increasing while the detected bit accuracy decreases.}
  \label{fig:fine-tune}
    \end{minipage}
\end{figure}

\noindent{\textbf{Threat via neural compression.}} Malicious users may alter the image to remove the watermark with deep learning techniques, like methods used for image compression with neural auto-encoders~\cite{cheng2020learned, minnen2018joint,balle2018variational_bmsh2018}. 
We evaluate the robustness of the watermark against neural auto-encoders at different compression rates (\cref{fig:image_level_attack}).  Specifically, when the bit accuracy decreases to $50\%$, a threshold indicating watermarking failure, the image quality significantly deteriorates. This suggests that our watermarking method is resilient to the effects of auto-encoder models. Even in extreme cases where bit accuracy reaches $50\%$, the image quality is so compromised that the images are no longer suitable for sharing or distribution purposes.

\noindent{\textbf{Threat via watermarking base model (white-box attack). }} 
In this scenario, attackers obtain the watermarking base model used by the NeRF creator. Then, malicious users can implement adversarial attacks~\cite{madry2017towards} with a leaked watermarking base model to remove copyright messages in rendered images. We launch a Projected gradient descent (PGD) attack~\cite{madry2017towards}, a classic adversarial attack on rendered images. As depicted in \cref{fig:image_level_attack}, experimental results show attackers can remove messages with minimal distortion to the rendered content. This highlights the importance of keeping the watermarking base models confidential~\cite{li2024graph,fernandez2023stable}.


\noindent{\textbf{Threat on the representation of NeRF.}} 
If malicious users have obtained NeRF models, they may choose to eliminate the watermarks by finetuning the models with no-watermark images. This process involves finetuning NeRF only via the content loss in \cref{eq:recon} without the message embedding process. 

\begin{figure}[t]
  \centering
\includegraphics[width=0.75\linewidth]{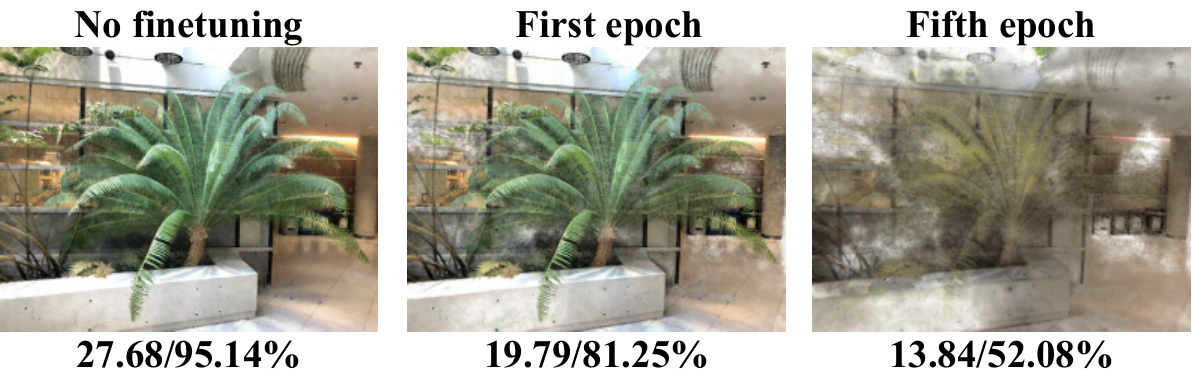}
    \vspace{-0.2cm}
   \caption{Attackers might fine-tune watermarked NeRF with unrelated scenes to remove the watermarks. However, the model quality significantly decreases after fine-tuning NeRF via unrelated images. The text below the image shows the results of PSNR and bit accuracy.} 
   \vspace{-0.5cm}
   \label{fig:fine-tune2}
\end{figure}

We assume that malicious users can obtain the original images corresponding to the scenes represented by NeRF. Then, from the results shown in \cref{fig:fine-tune}, the bit accuracy indeed drops, highlighting the need to securely manage both the original data and the NeRF models. However, if malicious users cannot obtain the images originally used for creating the NeRF, launching the attack becomes difficult. For example, if they directly use some images unrelated to the scenes represented by NeRF, it may significantly change the information stored in NeRF, making the model sharing unfeasible (as shown in \cref{fig:fine-tune2}). 

\section{Conclusion}
\label{sec:conclusion}
This paper introduces a plug-and-play method for watermarking NeRF during their creation. Utilizing a base watermarking model, NeRF creators can embed binary messages directly into their models as they are being developed. We propose a progressive global rendering to integrate message patterns into NeRF. The base model inherently offers basic robustness to its users. Our experimental results showcase the effectiveness of our approach. Our base model is compatible with NeRF variants, offering a more versatile watermarking solution.

\noindent{\textbf{Limitations.}} Though we have provided an effective technical solution for protecting the copyright of NeRF, copyright protection should be a multi-pronged effort involving various stakeholders. The legitimate enforcement of copyright requires a comprehensive strategy that goes beyond technological solutions. Legislative measures from relevant parties are also important to establish a regulatory framework that safeguards intellectual property rights.

\noindent \textbf{Acknowledgements.} Renjie's Research Group is supported by the National Natural Science Foundation of China under Grant No. 62302415, Guangdong Basic and Applied Basic Research Foundation under Grant No. 2022A1515110692, 2024A1515012822, and the Blue Sky Research Fund of HKBU under Grant No. BSRF/21-22/16.

\bibliographystyle{splncs04}
\bibliography{reference}

\end{document}